\documentclass[10pt,letterpaper]{article}
\usepackage[margin=1.01in]{geometry} 
\usepackage{graphicx}
\usepackage{subcaption}
\usepackage{amsmath}
\usepackage{multirow}
\usepackage{amsthm}
\usepackage{natbib}
\usepackage{hyperref}
\usepackage{amsfonts}
\usepackage{amssymb}
\usepackage{algorithm}
\usepackage{algorithmic}
\usepackage{mathtools}
\usepackage[most]{tcolorbox}
\usepackage{xcolor}
\usepackage{booktabs}   
\usepackage{caption}    
\usepackage{float}      
\usepackage{pifont}
\usepackage{authblk}

\title{LMT: A Bayesian Framework for Causal Discovery from Textual Alarm Records in Manufacturing Systems}
\author[1]{Xiaofeng Xiao}
\author[1]{Jianhong Chen}
\author[2]{Qiuzhuang Sun}
\author[3]{Naichen Shi}
\author[1]{Xubo Yue\thanks{Corresponding Author: \texttt{x.yue@northeastern.edu}}}
\affil[1]{Department of Mechanical \& Industrial Engineering, Northeastern University, Boston, MA, USA}
\affil[2]{College of Integrative Studies, Singapore Management University, Singapore}
\affil[3]{Department of Industrial Engineering and Management Sciences, Department of Mechanical Engineering, Northwestern University, IL, USA}

\begin{document}

\maketitle
\begin{abstract}
Textual event records, such as alarm logs, have become an increasingly common data source in engineering and manufacturing systems. Beyond identifying correlations or recurring patterns, engineers are often interested in understanding which types of events causally trigger or influence other events during system operation. Textual event descriptions may contain semantic clues about such causal relationships, and recent large language models (LLMs) provide a promising tool for extracting these signals. However, relying solely on LLM-encoded textual information is insufficient for accurate causal discovery, since semantic patterns do not directly reveal causal mechanisms and may confuse causation with correlation or frequent sequential patterns. To address these challenges, we propose \textbf{LMT}, a Bayesian causal discovery framework for engineering event data that jointly leverages textual descriptions and timestamps. Specifically, LMT first uses LLMs to extract semantic causal signals from event descriptions and constructs a prior distribution over causal graphs among event types or event clusters. It then incorporates temporal evidence through a Poisson-process-based likelihood, allowing the LLM-informed prior to be refined by timestamp-based statistical evidence. By integrating the textual and temporal information, LMT produces a causal graph that is both interpretable and data-supported. Simulation studies show that the proposed framework is effective across different settings and is especially advantageous in small-sample alarm-event scenarios.
\end{abstract}

\section{Introduction}

Manufacturing systems increasingly record data in textual form. These texts often document events occurring during manufacturing processes, such as alarms, maintenance actions, and process abnormalities \citep{geigle2018feature,ansari2020cost,cajner2026manufacturing}. Extracting useful information from such textual data of event description can help improve the reliability and quality of manufacturing systems. For example, event descriptions can provide valuable signals for predictive maintenance \citep{wang2017predictive}, and embeddings derived from event descriptions \citep{bhardwaj2021custom} can be effectively incorporated into deep learning models for manufacturing image analysis \citep{chen2023text,kwon2025predictive}. These analyses of event dataset often focus on identifying patterns, similarities or statistical associations among events. 
While such correlation-based information is useful for monitoring and prediction, it does not directly explain how the effect of events propagate through an engineering system. Thus, in addition to identifying correlations in event logs, engineers are often interested in a deeper question: \textit{which types of events causally trigger or influence other types of events?} Such causal relationships play an important role in understanding and managing complex engineering systems \citep{zhou2021end, ma2023nonlinear,vukovic2022causal,nadim2023data, xiao2025causality, chen2025federated, xiao2025bayesian, cheng2025survey}. 


Recently, large language models (LLMs) have shown strong capabilities in understanding and reasoning over textual information \citep{de2024impact,chen2024systems}. Since event descriptions may contain semantic clues about inter-event causality, LLMs provide a natural tool for extracting such information from textual event data 
\citep{chen2024large,li2025event, liu2025eventgpt}. This creates new opportunities for analyzing causal relationships in text-based event records from engineering systems.

\textit{However, relying only on LLM-encoded textual information is insufficient for accurate causal discovery.} Although LLMs can capture rich semantic patterns from event descriptions, they do not directly identify causal mechanisms \citep{ban2025llm,wu2025llm}. In particular, language models may confuse causation with correlation or frequent sequential patterns in text. This limitation is rooted in how LLMs are trained: they learn statistical associations among tokens and phrases across large corpora, rather than explicit interventional, temporal, or physical mechanisms that distinguish cause from effect \citep{pearl2009causal}. Thus, two event types that frequently appear together or are described in a fixed narrative order may receive a strong directional association, even when the relation reflects textual convention rather than physical causation.  Moreover, the interpretability of LLMs remains limited for characterizing causal relations in complex engineering systems, where event dependencies are often governed by underlying physical and temporal mechanisms rather than textual semantics alone. Thus, \textit{LLM-informed textual semantics should be treated as a useful but imperfect initial source, which must be further refined by more interpretable and verifiable methods.}


This paper proposes \textbf{LMT}, a language-model-prior guided Bayesian causal discovery framework for identifying causal relationships among event data in engineering systems by jointly utilizing timestamps and textual descriptions. Specifically, we first use LLMs to extract semantic causal signals from event descriptions. Based on these LLM-derived semantic signals, we construct a prior distribution over directed acyclic graphs (DAGs), where each node represents an event type or event cluster, and each directed edge represents a potential causal relationship between two event types. Then, to avoid relying solely on the LLM-based prior, we introduce a Poisson-process-based likelihood to incorporate temporal evidence from event timestamps. Under this likelihood, each event type is modeled through its occurrence intensity over time \citep{arnold2007temporal, finkbeiner2024synthesis, gong2024causal}. A candidate causal edge is encouraged when past occurrences of the source event help explain the future occurrence tendency of the target event. Finally, the LLM-informed prior is integrated with the Poisson temporal likelihood within a Bayesian posterior formulation. The posterior distribution integrates semantic knowledge from event descriptions and statistical evidence from event timestamps, producing a causal graph that is both interpretable and data-supported. The learned DAG can help reveal how engineering events propagate through a system, providing useful insights in complex industrial environments. 

Our contributions are summarized as follows:
\begin{itemize}
    \item We propose a Bayesian causal discovery framework for engineering event data that jointly utilizes textual descriptions and timestamps. Instead of relying on a single source of information, the proposed framework assigns textual and temporal information to different probabilistic roles. This design improves the reliability and usability of LLM-derived information in manufacturing systems.
 
    \item The proposed framework infers an interpretable causal graph (DAGs) among engineering event types or event clusters. The learned graph can provide useful insights for causal discovery, root-cause analysis, and decision-making in complex engineering systems.

    \item The proposed method is particularly suitable for alarm event data, where only limited event records are typically available in practice because engineering systems are often stopped and diagnosed shortly after alarms are triggered. This makes the framework especially useful for causal discovery in small-sample engineering settings. 
\end{itemize}

The rest of this paper is organized as follows. Section~\ref{sec:lr} reviews the related literature. Section~\ref{sec:pre} introduces the preliminary methods. Section~\ref{sec:method} presents the details of the proposed framework. Section~\ref{sec:simula} reports the simulation studies. Finally, we discuss potential improvements and conclude the paper. 
The source code for LMT is available at: \url{https://github.com/xx987/LMT}.

\section{Literature Review}\label{sec:lr}
\paragraph{LLM-driven Causal Discovery} LLMs have become powerful tools across many fields. As language models trained on large-scale textual data, they can provide rich semantic information and increasingly accurate contextual reasoning. A growing body of work has proposed using LLMs to augment causal discovery. \cite{kiciman2023causal} discussed that LLMs are able to generate content involving real causal relationships. Since causal statements and textual patterns may already exist in the training corpus, it is reasonable to consider LLMs as a potential source of prior knowledge for causal discovery. \cite{vashishtha2023causal} described that LLMs are not only useful for reasoning about causal graphs, but can also be used to explore causal effects after discovery, since LLMs may estimate effects when provided with sufficiently informative prompts. Similarly, \cite{long2023causal} utilized LLMs as imperfect experts for constructing priors in causal discovery, and then used a greedy search approach to recover the causal graph from the Markov equivalence class (MEC) based on LLM-provided information. Later, \cite{ban2025llm} incorporated constraints into the loss function to control the errors introduced by LLMs in causal reasoning. \cite{wu2025llm} further emphasized that causal information provided by LLMs should be constrained and should not be treated as the final decision; instead, scoring functions or information entropy can be used to restrict errors in LLM-derived priors. Most recently, \cite{zanna2026uncovering} demonstrated how different prompting strategies can guide LLMs to perform active learning in causal discovery and distinguish correlations from causal relationships more precisely. \cite{li2026mitigating} proposed that hallucination in LLMs can introduce bias into causal discovery, and that such hallucination bias can also be modeled within a causal graph. Therefore, \textit{LLMs can serve as efficient but imperfect experts} for providing initial signals about causal relationships in many types of datasets, including engineering event data. However, \textit{the prior information generated by LLMs still needs to be calibrated and adjusted using additional data-driven evidence}. 

\paragraph{Event Data Causality}Event data are inherently associated with temporal signals; therefore, causal discovery from event datasets often incorporates temporal information \citep{mirza2021event, gong2024causal,mbogu2024data}. \cite{tang2018nodemerge} designed an event data storage method based on temporal generation, which makes event data more convenient for causal analysis. Similarly, \cite{zuo2021learnda} proposed an approach for event data imputation by using temporal generation methods to create additional data for causal discovery. \cite{rebboud2022beyond} modeled event datasets using representation learning methods and then mapped events into a causal graph to learn causal relationships among events.

In addition, many studies have used temporal point processes, especially Poisson-process-based models, to learn event causality from temporal information. \cite{jin2020visual} applied the Hawkes process, a self-exciting point process closely related to Poisson processes, to causal discovery in visual event sequences. In this setting, visual sequences are encoded as event-generation processes, and root causes can be identified by modeling temporal dependencies among events through the point process.  \cite{cuppers2024causal} introduced an information-entropy function to measure the cost of greedy search in Hawkes-process-based causal discovery, thereby constraining the search space of causal graphs inferred from point-process models. \cite{liu2024tnpar} mapped events generated from a Poisson process into a neural network and investigated how the network parameters can represent causal relationships. Similarly, \cite{chang2025anhp} proposed that neural point processes can learn the structure of causal graphs and provide causal information for event datasets. Most recently, \cite{li2025poisson} utilized Poisson processes to decompose information in event data, and the decomposed information was further evaluated using different machine learning methods to identify potential causal graphs. 

Based on these studies, we introduce the Hawkes process as an adjustment tool for the LLM-derived prior, since Hawkes-process-based models have demonstrated strong capability in modeling temporal dependencies in event datasets. Specifically, LLM-based experts provide initial prior information from the textual descriptions of events, while the Hawkes process incorporates temporal signals through a likelihood function to facilitate the inference of a more accurate posterior causal graph.

\section{Preliminaries}\label{sec:pre}

\subsection{Data Setting}

We consider an engineering system that continuously generates event records over a finite observation time interval $[0,T]$. 
Formally, the raw event dataset is denoted as
$
\mathcal{D}_{\mathrm{raw}}
=
\{(t_i,s_i)\}_{i=1}^{n}
$
where $i$ is the index of an observed event, $n$ is the total number of observed events, $t_i \in [0,T]$ is the timestamp of event $i$, and $s_i$ is the raw textual description associated with event $i$. The timestamps describe when events occur, while the textual descriptions provide semantic information about what happens in the engineering system.

Based on this raw event dataset, we further extract text embeddings and do clustering for these events which will be used for downstream causal discovery. 
We use a Hugging Face Sentence-Transformer model for embedding the alarm texts \citep{jain2022hugging}. This provides a straightforward way to cluster events based on the similarity of their sentence embeddings. The underlying assumption is that alarm records with similar textual descriptions are likely to correspond to similar event types or functional modules. Therefore, the resulting text-based clusters provide an initial event grouping for subsequent causal discovery; more details are provided in Sec.~\ref{sec:simula}.  After text embedding and clustering, each event is represented as $e_i = (t_i, k_i, x_i)$, $i=1,\ldots,n$ where $k_i \in \{1,\ldots,K\}$ is the semantic cluster label of event $i$, $K$ is the total number of semantic clusters, and $x_i \in \mathbb{R}^d$ is the embedding vector of the textual description $s_i$. Therefore, the observed event dataset is denoted as $\mathcal{D}=
\{(t_i,k_i,x_i)\}_{i=1}^{n}$. Here, $\mathcal{D}$ contains both temporal information and semantic information. The timestamps $\{t_i\}_{i=1}^{n}$ describe when events occur, while the cluster labels $\{k_i\}_{i=1}^{n}$ and embeddings $\{x_i\}_{i=1}^{n}$ describe what types of events occur and what semantic meanings they carry.

\subsection{DAG Prior}

The objective of this work is to infer a causal structure among event clusters. Since each cluster represents a semantic type of engineering event, we model the causal relationships among clusters by a DAG \citep{lorch2021dibs}. Let $G = (\mathcal{V}, \mathcal{E})$ where $\mathcal{V}=\{1,\ldots,K\}$ denotes the set of $K$ event clusters, and $\mathcal{E}$ denotes the set of directed edges. In this work, this prior is constructed from LLM-derived semantic information. The key motivation is that event descriptions often contain useful domain-level causal clues. For example, some descriptions may indicate upstream fault conditions, while others may indicate downstream system responses. Therefore, LLMs can be used as imperfect but informative experts to provide initial beliefs about which cluster-level causal relations are more plausible. Based on these LLM-derived information, we define a graph prior over the adjacency matrix as $p(G \mid E)$
where $E$ denotes the textual information of event clusters. This prior assigns higher probability to graph structures that are more consistent with the causal knowledge extracted from event descriptions by the LLM.



\section{Methodology}\label{sec:method}
\subsection{DAG Prior from LLM-based Imperfect Expert Embeddings}\label{sec:4.1}

To incorporate textual information into the causal prior, we first use a LLM to extract role-specific semantic scores for each event. 
For an event $e_i$ with textual description $s_i$, the LLM is asked to evaluate the description only and assign two independent continuous scores:
\[
c_i^{\mathrm{LLM}} \in [0,1], 
\qquad 
r_i^{\mathrm{LLM}} \in [0,1]
\]
where $c_i^{\mathrm{LLM}}$ denotes the degree to which the event text suggests an upstream causal or triggering role, and $r_i^{\mathrm{LLM}}$ denotes the degree to which the event text suggests a downstream effect or symptom role. 
Importantly, the LLM is not provided with temporal information, event labels, graph structure, or any model output. 
Therefore, these scores should be interpreted as semantic prior information extracted from the textual descriptions rather than causal conclusions inferred from the event sequence.

The initial prompt example used to query the LLM is summarized in Box~\ref{box:llm_prompt}.

\begin{tcolorbox}[
    colback=black!5,
    colframe=gray!50,
    title={Excerpt of LLM Prompt for Event Role Scoring},
    label={box:llm_prompt},
    fonttitle=\bfseries\small,
    fontupper=\footnotesize,
    boxsep=1.5pt,
    left=3pt,
    right=3pt,
    top=3pt,
    bottom=3pt,
    arc=1pt,
    boxrule=0.4pt,
    before skip=4pt,
    after skip=4pt,
    breakable
]
You are labeling industrial alarm events with the alarm text provided.

For each alarm text, output two independent continuous scores in $[0,1]$:

\vspace{-0.4em}
\begin{itemize}
    \setlength\itemsep{0pt}
    \setlength\parsep{0pt}
    \setlength\topsep{0pt}
    \item \texttt{cause\_like}: the degree to which the wording suggests an upstream trigger, initiating factor, or root-cause role.
    \item \texttt{effect\_like}: the degree to which the wording suggests a downstream symptom, consequence, or affected-system role
\end{itemize}
\vspace{-0.6em}

\[
\cdots
\]
\end{tcolorbox}

The LLM-generated scores are used to supervise the role-specific projections $W_c$ and $W_e$, which map the original text embedding into cause-role and effect-role representation spaces \citep{xie2019distributed}. 
Specifically, for each event $i$, we define $h_i^{(c)} = W_c x_i$, $h_i^{(e)} = W_e x_i$, 
where recall $x_i$ denotes the base text embedding of event $i$, while $h_i^{(c)}$ and $h_i^{(e)}$ denote its cause-role and effect-role representations, respectively. Here, the embeddings $x_i$ are obtained from a separate event dataset used only for training the causal embedding model. This dataset is independent of the real timestamped event dataset used later in the temporal likelihood.
Two lightweight prediction heads are then used to regress the LLM-provided role scores as: $\hat c_i = \sigma\!\left(u_c^\top h_i^{(c)} + b_c\right)$ and
$\hat e_i = \sigma\!\left(u_e^\top h_i^{(e)} + b_e\right)
$, where $u_c,u_e$ are learnable vectors, $b_c,b_e$ are learnable scalar biases, and $\sigma(\cdot)$ is the sigmoid function. 
The role-distillation loss is defined as
$$
\mathcal{L}_{\mathrm{role}}
=
\sum_i
\left[
\left(\hat c_i - c_i^{\mathrm{LLM}}\right)^2
+
\left(\hat e_i - e_i^{\mathrm{LLM}}\right)^2
\right].
$$
Minimizing this objective encourages $W_c$ to extract cause-like semantic features and $W_e$ to extract effect-like semantic features from the event text embeddings.

After obtaining the role-specific representations, we define a directional text-based cause--effect matching score for an event pair $i\to j$:
$$
a_{ij}
=
\left(h_i^{(c)}\right)^\top h_j^{(e)}
=
\left(W_c x_i\right)^\top \left(W_e x_j\right).
$$
Here, $a_{ij}$ measures the semantic compatibility between event $i$ as a potential cause and event $j$ as a potential effect. 
A larger value of $a_{ij}$ indicates that the text of event $i$ is more consistent with a cause-like role, the text of event $j$ is more consistent with an effect-like role, and the directed pair $i\to j$ is more plausible from the perspective of text-based role semantics.

We then normalize this score using a sigmoid function:
$$
r_{ij}^{\mathrm{text}}
=
\sigma(a_{ij})
=
\sigma\!\left(
\left(W_c x_i\right)^\top
\left(W_e x_j\right)
\right)
$$
where $r_{ij}^{\mathrm{text}}\in(0,1)$ represents the probability that  event $i$ is as a potential cause and event $j$ is as a potential effect based on their textual semantics. 
Notably, this score is not treated as a true causal probability; instead, it is used as a directional semantic prior score for the candidate edge $i\to j$.

To obtain a cluster-level causal prior, we aggregate the event-level directional semantic scores according to their cluster labels. Let $\mathcal{I}_a=\{i:k_i=a\}$ denote the index set of events belonging to cluster $a$. For two clusters $a$ and $b$, we define the cluster-level textual causal score by averaging all event-level scores from cluster $a$ to cluster $b$:
$$
q_{ab}^{\mathrm{LLM}}
=
\frac{1}{|\mathcal{I}_a||\mathcal{I}_b|}
\sum_{i\in \mathcal{I}_a}
\sum_{j\in \mathcal{I}_b}
r_{ij}^{\mathrm{text}}.
$$
Here, $q_{ab}^{\mathrm{LLM}}\in(0,1)$ represents the LLM-informed prior belief that cluster $a$ is a potential cause of cluster $b$. A larger value of $q_{ab}^{\mathrm{LLM}}$ means that, on average, events in cluster $a$ are more semantically likely to act as causes of events in cluster $b$. For self-edges, we set
$q_{aa}^{\mathrm{LLM}}=0$ since self-loops are excluded in the cluster-level DAG. 

We then denote the continuous adjacency matrix of the cluster-level causal graph as $G=[G_{ab}]\in(0,1)^{K\times K}$ where $G_{ab}$ represents the soft edge strength, or relaxed edge probability, from cluster $a$ to cluster $b$. 
A larger value of $G_{ab}$ indicates stronger evidence that cluster $a$ causally influences cluster $b$, while a smaller value indicates weaker evidence. 
Instead of sampling a discrete graph from the LLM-derived scores, we define a continuous prior density over the soft adjacency matrix $G$. 
The LLM-derived score $q_{ab}^{\mathrm{LLM}}$ is used as a prior preference for the relaxed edge strength $G_{ab}$. However, the LLM-informed graph prior alone does not guarantee that the learned graph is acyclic. 
Since the target cluster-level causal structure is a DAG, we further introduce a differentiable acyclicity constraint \citep{zheng2018dags}. 
Following the continuous DAG characterization, we define $h(G)=\operatorname{tr}\left(\exp(G\odot G)\right)-K$
where $\odot$ denotes the element-wise product and $K$ is the number of clusters. 
This function satisfies $h(G)=0$ if and only if the weighted directed graph induced by $G$ is acyclic. 
When $G$ contains directed cycles, $h(G)$ becomes positive and increases with the strength of cyclic structure \citep{lorch2021dibs}.

Thus, the LLM-informed DAG prior is defined as
$$
p(G\mid E_{\mathrm{LLM}})
\propto
\exp\left\{
\lambda_{\mathrm{LLM}}
\sum_{a\neq b}
\left[
G_{ab}\log q_{ab}^{\mathrm{LLM}}
+
(1-G_{ab})\log\left(1-q_{ab}^{\mathrm{LLM}}\right)
\right]
-
\beta_{\mathrm{dag}} h(G)
\right\}
$$
where $\beta_{\mathrm{dag}}>0$ controls the strength of the acyclicity penalty, $E_{\mathrm{LLM}}$ denotes the textual information used by the LLM, and $\lambda_{\mathrm{LLM}}>0$ controls the strength of the LLM-informed prior.

\subsection{Refining the Prior with Temporal Likelihood}
So far, the information extracted from the event dataset is based only on the textual descriptions of the events. 
As specified in the prompt, the LLM is given only the textual description of each event.
Similarly, the embedding training process also relies only on the event text. 
Therefore, the resulting textual prior alone is insufficient to fully explain causal relations among events, especially when the textual descriptions are ambiguous or repetitive. For example, in an engineering system, two events may have very similar or even identical descriptions, such as ``temperature alarm triggered''. From text alone, it is difficult to determine whether one event caused the other. Therefore, temporal information plays a crucial role in refining the text-based prior. By incorporating the temporal event sequence through a likelihood function, we can update the prior belief and obtain the posterior distribution of the causal graph. 

To incorporate temporal dependencies among event clusters, we introduce a Hawkes-process likelihood. 
A Hawkes process is a self- and mutually-exciting point process that models how past events increase the occurrence intensity of future events. 
It has been widely used to model temporal triggering effects and event-level Granger causality \citep{xu2016learning,salehi2019learning,cai2022thps}. 
In our setting, the LLM-informed prior provides a text-based belief over the cluster-level causal graph, while the temporal likelihood evaluates whether the observed event sequence is consistent with a candidate soft graph $G$.

Let the observed temporal event sequence be $\mathcal{D}_{\mathrm{time}}=\{(t_n,k_n)\}_{n=1}^{N}$ where $t_n$ is the timestamp of event $n$, and $k_n\in\{1,\dots,K\}$ denotes its cluster label. 
The events are ordered such that $t_1<t_2<\cdots<t_N$. 
For each cluster $b\in\{1,\dots,K\}$, we define the conditional intensity function as
$$
\lambda_b(t \mid \mathcal{H}_t, G, \Theta)
=
\mu_b
+
\sum_{m:\, t_m<t}
G_{k_m,b}\,\kappa_{k_m,b}(t-t_m)
$$
where $\mathcal{H}_t$ denotes the event history before time $t$, and $\Theta=\{\mu_b,\beta_{ab}:a,b=1,\dots,K\}$ collects the temporal parameters. 
Here, $\lambda_b(t \mid \mathcal{H}_t, G, \Theta)$ is the instantaneous occurrence rate of an event from cluster $b$ at time $t$. 
The parameter $\mu_b>0$ is the background intensity of cluster $b$, representing spontaneous occurrences that are not triggered by previous events. 
For a past event $m$ with cluster label $k_m$, the term $G_{k_m,b}$ represents the relaxed causal edge strength from cluster $k_m$ to cluster $b$, and $\kappa_{k_m,b}(t-t_m)$ controls how the influence decays over time.

In this work, we use the exponential kernel $\kappa_{ab}(\tau)=\beta_{ab}e^{-\beta_{ab}\tau}\mathbf{1}[\tau>0]$
where $\tau$ is the time lag, $\beta_{ab}>0$ is the decay rate for the directed pair $a\to b$, and $\mathbf{1}[\tau>0]$ ensures that only past events can influence future events. 
Substituting this kernel into the intensity function gives
$$
\lambda_b(t \mid \mathcal{H}_t, G, \Theta)
=
\mu_b
+
\sum_{m:\, t_m<t}
G_{k_m,b}\,\beta_{k_m,b}
e^{-\beta_{k_m,b}(t-t_m)}.
$$
Given the observed sequence $\mathcal{D}_{\mathrm{time}}$, the likelihood under the multivariate Hawkes process is \citep{cui2023robust}:
$$
p(\mathcal{D}_{\mathrm{time}} \mid G,\Theta)
=
\left[
\prod_{n=1}^{N}
\lambda_{k_n}(t_n \mid \mathcal{H}_{t_n},G,\Theta)
\right]
\exp\!\left(
-\int_0^T
\sum_{b=1}^{K}
\lambda_b(t \mid \mathcal{H}_t,G,\Theta)\,dt
\right).
$$
The product term encourages the model to assign high intensity to the observed events at their actual occurrence times, while the compensator term prevents the model from assigning excessively large intensity over the entire observation window. Taking logarithms, the temporal log-likelihood is
$$
\mathcal{L}_{\mathrm{time}}(G,\Theta)
=
\sum_{n=1}^{N}
\log
\lambda_{k_n}(t_n \mid \mathcal{H}_{t_n},G,\Theta)
-
\int_0^T
\sum_{b=1}^{K}
\lambda_b(t \mid \mathcal{H}_t,G,\Theta)\,dt.
$$

Under the exponential kernel, the intensity evaluated at an observed event time $t_n$ can be written explicitly as
$$
\lambda_{k_n}(t_n \mid \mathcal{H}_{t_n},G,\Theta)
=
\mu_{k_n}
+
\sum_{m<n}
G_{k_m,k_n}\,\beta_{k_m,k_n}
e^{-\beta_{k_m,k_n}(t_n-t_m)}.
$$
Here, the condition $m<n$ is valid because the events are ordered by time. The integral term also has a closed-form expression:
$$
\int_0^T
\lambda_b(t \mid \mathcal{H}_t,G,\Theta)\,dt
=
\mu_b T
+
\sum_{m=1}^{N}
G_{k_m,b}\,
\left(
1-e^{-\beta_{k_m,b}(T-t_m)}
\right).
$$
Therefore, the full temporal log-likelihood can be written as
$$
\begin{aligned}
\mathcal{L}_{\mathrm{time}}(G,\Theta)
=&
\sum_{n=1}^{N}
\log\!\left(
\mu_{k_n}
+
\sum_{m<n}
G_{k_m,k_n}\,\beta_{k_m,k_n}
e^{-\beta_{k_m,k_n}(t_n-t_m)}
\right) -
\sum_{b=1}^{K}\mu_b T \\
&-
\sum_{b=1}^{K}
\sum_{m=1}^{N}
G_{k_m,b}\,
\left(
1-e^{-\beta_{k_m,b}(T-t_m)}
\right).
\end{aligned}
$$ Combining the temporal likelihood with the LLM-informed DAG prior gives the posterior distribution
$$
p(G,\Theta \mid \mathcal{D}_{\mathrm{time}},E_{\mathrm{LLM}})
\propto
p(\mathcal{D}_{\mathrm{time}}\mid G,\Theta)
p(G\mid E_{\mathrm{LLM}})
$$
where $p(G\mid E_{\mathrm{LLM}})$ is the continuous LLM-informed DAG prior defined in the previous subsection.
Equivalently, the MAP objective can be written as

\begin{equation}\label{eq:optimizer}
(G^\star,\Theta^\star)
=
\arg\max_{G,\Theta}
\left[
\mathcal{L}_{\mathrm{time}}(G,\Theta)
+
\log p(G\mid E_{\mathrm{LLM}})
\right].
\end{equation}

The MAP objective in Eq.~\ref{eq:optimizer} is optimized through a continuous relaxation. 
As mentioned in Sec.~\ref{sec:4.1}, the cluster-level graph is represented by a soft adjacency matrix $G=[G_{ab}]\in(0,1)^{K\times K}$
with $G_{aa}=0$ to exclude self-loops. 
To optimize $G$ while keeping each edge strength in $(0,1)$, we introduce an unconstrained edge-logit matrix $S=[S_{ab}]\in\mathbb{R}^{K\times K}$ and parameterize $G_{ab}=\sigma(S_{ab})$, $a\neq b$
where $\sigma(\cdot)$ is the sigmoid function. 
Thus, $S_{ab}$ is the unconstrained optimization variable, while $G_{ab}$ is the corresponding relaxed edge strength used in the likelihood and prior.

The Hawkes parameters are also optimized through unconstrained variables. 
Specifically, we introduce $\tilde{\mu}_b\in\mathbb{R}$ and $\tilde{\beta}_{ab}\in\mathbb{R}$ and set
$\mu_b=\mathrm{softplus}(\tilde{\mu}_b)$, $\beta_{ab}=\mathrm{softplus}(\tilde{\beta}_{ab})$ which ensures $\mu_b>0$ and $\beta_{ab}>0$. 
Let $\tilde{\Theta}$ denote the collection of these unconstrained Hawkes parameters:
$\tilde{\Theta}=\{\tilde{\mu}_b,\tilde{\beta}_{ab}:a,b=1,\dots,K\}$.



Under these parameterizations, both the Hawkes log-likelihood and the LLM-informed graph prior are differentiable with respect to the unconstrained parameters $S$ and $\tilde{\Theta}$. 
Therefore, instead of directly optimizing the constrained variables $G$ and $\Theta$, we optimize the following negative MAP loss:
$$
\mathcal{J}(S,\tilde{\Theta})
=
-
\mathcal{L}_{\mathrm{time}}(G,\Theta)
-
\log p(G\mid E_{\mathrm{LLM}}).
$$ 
The optimization variables are $S$ and $\tilde{\Theta}$, while $G$ and $\Theta$ are transformations of these variables.

At each iteration $r$, we first compute
$G^{(r)}=\sigma(S^{(r)})$, $\mu_b^{(r)}=\mathrm{softplus}(\tilde{\mu}_b^{(r)})$, $\beta_{ab}^{(r)}=\mathrm{softplus}(\tilde{\beta}_{ab}^{(r)})$. 
Then we evaluate the negative MAP loss $\mathcal{J}(S^{(r)},\tilde{\Theta}^{(r)})$ using the Hawkes temporal log-likelihood and the LLM-informed graph prior. 
The gradients are computed by automatic differentiation:
$$
\nabla_S \mathcal{J}
=
-
\nabla_S \mathcal{L}_{\mathrm{time}}(G,\Theta)
-
\nabla_S \log p(G\mid E_{\mathrm{LLM}}),
$$
$$
\nabla_{\tilde{\Theta}} \mathcal{J}
=
-
\nabla_{\tilde{\Theta}} \mathcal{L}_{\mathrm{time}}(G,\Theta)
$$
where the graph prior does not depend directly on $\tilde{\Theta}$.

The gradient with respect to $S$ is obtained through the chain rule:
$$
\frac{\partial \mathcal{J}}{\partial S_{ab}}
=
\frac{\partial \mathcal{J}}{\partial G_{ab}}
\frac{\partial G_{ab}}{\partial S_{ab}}
=
\frac{\partial \mathcal{J}}{\partial G_{ab}}
G_{ab}(1-G_{ab}).
$$
Similarly, the positivity-constrained Hawkes parameters are optimized through
$$
\frac{\partial \mathcal{J}}{\partial \tilde{\mu}_b}
=
\frac{\partial \mathcal{J}}{\partial \mu_b}
\frac{\partial \mu_b}{\partial \tilde{\mu}_b},
\qquad
\frac{\partial \mathcal{J}}{\partial \tilde{\beta}_{ab}}
=
\frac{\partial \mathcal{J}}{\partial \beta_{ab}}
\frac{\partial \beta_{ab}}{\partial \tilde{\beta}_{ab}}
$$
where $\frac{\partial \mu_b}{\partial \tilde{\mu}_b}
=\sigma(\tilde{\mu}_b)$, $\frac{\partial \beta_{ab}}{\partial \tilde{\beta}_{ab}}=\sigma(\tilde{\beta}_{ab})$ because the derivative of the softplus function is the sigmoid function.

Using standard gradient descent, the update rule can be written as
$$
S^{(r+1)}
=
S^{(r)}
-
\eta
\nabla_S \mathcal{J}(S^{(r)},\tilde{\Theta}^{(r)}),
$$
$$
\tilde{\Theta}^{(r+1)}
=
\tilde{\Theta}^{(r)}
-
\eta
\nabla_{\tilde{\Theta}} \mathcal{J}(S^{(r)},\tilde{\Theta}^{(r)})
$$
where $\eta>0$ is the learning rate. 
In implementation, we use Adam \citep{zhang2018improved,bock2019proof}, which replaces the raw gradients with adaptive first- and second-moment estimates, but optimizes the same negative MAP loss.  After optimization, the final binary graph is obtained by thresholding the learned soft adjacency matrix:
$\widehat{G}_{ab}=\mathbf{1}\{G_{ab}>\tau\}$, $a\neq b$ where $\tau$ is a selected threshold.

In summary, the temporal likelihood refines the text-based prior by checking whether the observed event sequence is temporally consistent with the candidate graph structure. 
The background intensity $\mu_b$ captures spontaneous occurrences in cluster $b$, while the triggering term captures how past events from cluster $a$ increase the future intensity of cluster $b$ through the soft edge strength $G_{ab}$, and temporal decay rate $\beta_{ab}$.

\section{Simulations}\label{sec:simula}

In this section, we evaluate our proposed method on a simulated chemical manufacturing event dataset. The objective is to infer causal relationships among alarm clusters from event texts. We compare our method with a couple benchmark approaches for event causality discovery and further conduct ablation studies to examine the contribution of each component.

\subsection{Simulated Dataset}

We first sample a ground-truth DAG over the $p$ nodes (clusters). Specifically, we draw a random topological ordering $\pi$ of the clusters. For each pair of distinct clusters $(i,j)$ satisfying $\pi(i)<\pi(j)$, we independently include the directed edge $i \to j$ with probability $\rho=0.5$; edges violating this ordering are not allowed. This construction guarantees that the sampled graph is acyclic. Given the DAG, each simulated event is assigned to one of the $p$ clusters and is associated with a textual alarm description generated by GPT-5 \citep{singh2025openai}. We then simulate event times with a multivariate Hawkes process from cluster \(i\) to cluster \(j\) is allowed in the ground-truth DAG. 

This event dataset basically contains three main components: the event identifier, the textual alarm description, and the alarm trigger time. We consider two simulation settings with different sample sizes: 500 events ($n=500$) and 1000 events ($n=1000$). 
An example layout of the dataset with first three event records is presented in Table~\ref{tab:simulated_data_example}. 

\begin{table}[H]
\centering
\caption{Example records from the simulated event dataset.}
\label{tab:simulated_data_example}
\renewcommand{\arraystretch}{1.15}
\setlength{\tabcolsep}{6pt}
\small
\begin{tabular}{c p{6.8cm} c}
\toprule
\textbf{Event\_id} & \textbf{Alarm Text} & \textbf{Time} \\
\midrule
1 &Feed pressure alarm triggered & 1/1/26 8:06:05 \\
2 & SIS recovery latch fault detected at state & 1/1/26 8:06:17 \\
3 & Catalyst feeder jitter alarm reported on line& 1/1/26 8:12:43 \\
\multicolumn{3}{c}{\(\cdots\)} \\
\bottomrule
\end{tabular}
\end{table}

We aim to identify the causal relationships among alarm clusters in these scenarios using our proposed method. Specifically, each alarm event is categorized into a cluster, and one cluster may serve as a potential cause or trigger of another cluster. Therefore, our goal is to infer a causal graph among these alarm clusters. Notably, the clustering strategy is scenario-dependent. Textual alarm events can be grouped in many different ways, and different clustering strategies may lead to different causal graphs and causal interpretations. Therefore, the choice of clustering method should depend on the specific dataset, system context, and research objective. In principle, even a predefined or random grouping strategy can be used if it is meaningful for the target scenario. In this paper, we cluster alarm events based on the similarity of their sentence embeddings of \textbf{Alarm Text} for ease of implementation. After clustering, the cluster identifier (i.e., the cluster index) of each event is associated with the original event dataset. We consider three scenarios about the numbers of clusters, corresponding to the nodes in the DAG, are $p=5$, $p=8$, and $p=16$.

\subsection{Baselines}

We compare LMT with three representative approaches for event causality identification.

\begin{itemize}
    \item \textbf{CASCADE} \citep{cuppers2024causal}: This method proposes a Minimum Description Length (MDL) framework for identifying cause--effect pairs from event sequences. It scores candidate causal relations based on description length and incorporates temporal information through Poisson-based event signals. Since it relies only on temporal information without using event text, it can be regarded as a temporal-only baseline for event causality discovery.
    
    \item \textbf{VOC-Emb} \citep{xie2019distributed}: In contrast to CASCADE, this method relies on textual information but does not incorporate temporal information. It represents event descriptions using embeddings learned from a dictionary of cause--effect pairs and word lists, and then derives causal scores from the resulting embedding space. Therefore, it serves as a text-only baseline.
    
    \item \textbf{LLM Prompting} \citep{hurst2024gpt}: Since LLMs have recently been widely applied across many domains, we also include a direct prompting baseline. In this setting, we ask an LLM to directly infer causal relations among event clusters from their textual descriptions associated with the temporal information, in order to evaluate whether a pure LLM-based approach can recover event causality without an explicit causal discovery framework.
\end{itemize} 

\subsection{Metrics}
We evaluate graph recovery performance using three standard metrics: Structural Hamming Distance (SHD), True Positive Rate (TPR), and False Discovery Rate (FDR). 

\paragraph{Structural Hamming Distance (SHD).} SHD measures the total number of edge mismatches between the estimated graph and the ground-truth graph. Each missing true edge and each extra incorrect edge contributes one unit to the distance. A lower SHD indicates better graph recovery performance, and $\mathrm{SHD}=0$ means that the estimated directed edge set exactly matches the ground-truth graph.

\paragraph{True Positive Rate (TPR).}
TPR, also referred to as recall, measures the fraction of true directed edges in the ground-truth graph that are correctly recovered by the estimated graph. A higher TPR indicates that fewer true edges are missed, i.e., the method produces fewer false negatives among the ground-truth edges.

\paragraph{False Discovery Rate (FDR).}
FDR measures the fraction of predicted directed edges that are incorrect, namely, edges that do not appear in the ground-truth graph. A lower FDR indicates that the estimated graph contains fewer spurious edges, i.e., fewer false positives among all predicted edges. In our setting, FDR is computed as $\mathrm{FDR}=\frac{\mathrm{FP}}{\mathrm{TP}+\mathrm{FP}}$ with the usual convention applied when no edges are predicted. Here, a true positive (TP) is a directed edge that is present in both the estimated graph and the ground-truth graph; 
a false positive (FP) is a directed edge that is predicted by the estimated graph but does not exist in the ground-truth graph; 
and a false negative (FN) is a directed edge that exists in the ground-truth graph but is missed by the estimated graph.

\subsection{Results}

The results are reported in Table~\ref{Tab:event500} and Table~\ref{Tab:event1000}. 
All results of LMT are averaged over 10 repetitions with different random seeds, and we report the mean and standard deviation. 
Overall, LMT achieves the best performance across different settings. 
In particular, LMT consistently obtains the lowest SHD and the highest TPR, while maintaining relatively low FDR compared with all baselines. 
This indicates that LMT is able to recover more true causal edges while introducing fewer spurious edges.

Under the setting of $N=500$ and $p=5$, LMT achieves near-perfect recovery, with SHD $=0.5\pm0.6$, TPR $=0.94\pm0.07$, and FDR $=0.16\pm0.08$, substantially outperforming the competing methods. 
As the problem becomes more challenging with larger numbers of clusters ($p=8$ and $p=16$), the performance of all methods degrades, but LMT remains clearly superior. 
A similar trend is observed when the number of events increases to $N=1000$, where LMT still achieves the best overall results across all three metrics.

These comparisons also highlight the limitations of the baselines. 
CASCADE, which relies only on temporal information, performs better than the text-only and pure-LLM baselines in some settings, but is still consistently worse than LMT. 
VOC-Emb, which uses only textual embeddings, and Pure-LLM, which directly infers causality from prompting, both show clear performance gaps, especially in terms of SHD and TPR. 
This suggests that neither temporal-only nor text-only modeling is sufficient for robust event causal discovery, whereas jointly leveraging complementary information leads to more accurate graph recovery.

\begin{figure}[H]
    \centering
      \scalebox{0.75}{
    \begin{subfigure}[b]{0.5\textwidth}
        \centering
        \includegraphics[width=\textwidth]{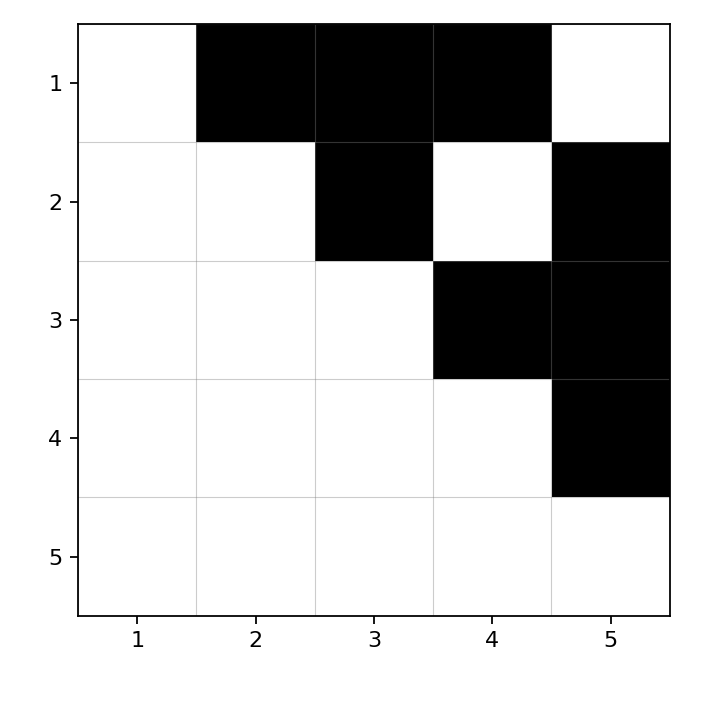}
        \caption{Ground-truth adjacency matrix}
        \label{fig:k5_500_event_gt}
    \end{subfigure}}
    \hfill
      \scalebox{0.75}{\begin{subfigure}[b]{0.5\textwidth}
        \centering
        \includegraphics[width=\textwidth]{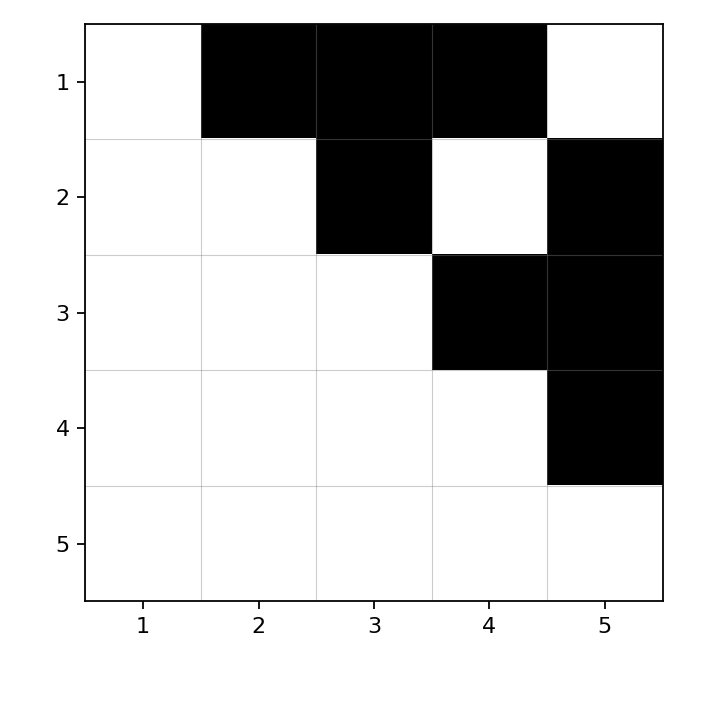}
        \caption{Estimated adjacency matrix by LMT}
        \label{fig:k5_500_event_ours}
    \end{subfigure}}
    \caption{Comparison between the ground-truth and inferred cluster-level directed adjacency matrices under the setting $p=5$ and $N=500$. Each nonzero entry represents a directed edge between two clusters. Black entries denote edges identified in the inferred graph, while \textcolor{red}{red} entries denote true edges missed by the inferred graph. This figure provides a qualitative visualization of graph recovery performance, showing that LMT recovers most of the true causal structure.}
    \label{fig:k5_500_event_compare}
\end{figure}

\begin{figure}[H]
    \centering
    \scalebox{0.75}{
    \begin{subfigure}[b]{0.5\textwidth}
        \centering
        \includegraphics[width=\textwidth]{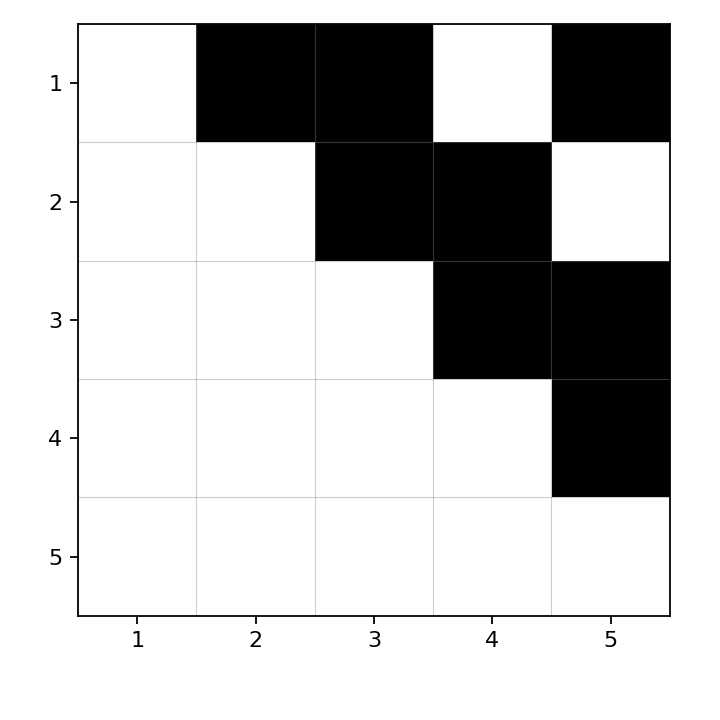}
        \caption{Ground-truth adjacency matrix}
        \label{fig:k5_1000_event_gt}
    \end{subfigure}}
    \hfill
    \scalebox{0.75}{
    \begin{subfigure}[b]{0.5\textwidth}
        \centering
        \includegraphics[width=\textwidth]{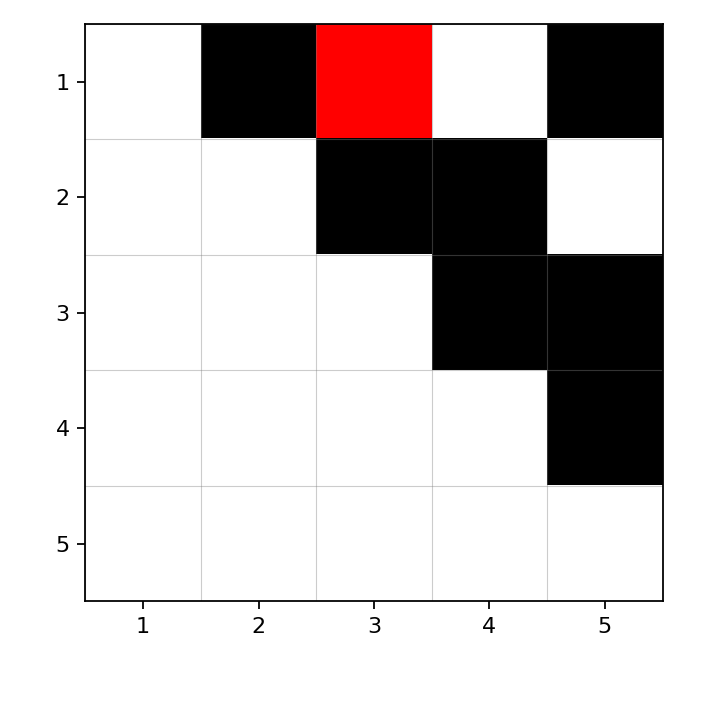}
        \caption{Estimated adjacency matrix by LMT}
        \label{fig:k5_1000_event_ours}
    \end{subfigure}}
    \caption{Comparison between the ground-truth and inferred cluster-level directed adjacency matrices under the setting $p=5$ and $N=1000$. Each nonzero entry represents a directed edge between two clusters. Black entries denote edges identified in the inferred graph, while \textcolor{red}{red} entries denote true edges missed or incorrectly estimated by the inferred graph. This figure provides a qualitative visualization of graph recovery performance, showing that LMT recovers most of the true causal structure.}
    \label{fig:k5_1000_event_compare}
\end{figure}

A qualitative visualization is shown in Figure~\ref{fig:k5_500_event_compare} and Figure~\ref{fig:k5_1000_event_compare}, where the inferred graph recovers most of the true causal edges and misses only one edge.

\begin{table}[H]
\centering
\scalebox{0.75}{
\begin{tabular}{c l c c c}
\toprule
\textbf{Metric} & \textbf{Method} & $p=5$ & $p=8$ & $p=16$ \\
\midrule

\multirow{5}{*}{\textbf{SHD}}
& \textbf{LMT (ours)}  & $0.5 \pm 0.6$ & $3.1 \pm 1.7$ & $7.6 \pm 2.6$ \\
& CASCADE  & $3.7 \pm 1.4$ & $8.2 \pm 2.2$ & $14.7 \pm 1.7$ \\
& VOC-Emb       & $10.6\pm 2.8$ & $ 14.6 \pm 0.9$ & $20.8 \pm 1.3$ \\
& Pure-LLM      & $7.8 \pm 0.4$ & $13.6 \pm 0.9$ & $19.6 \pm 5.8$ \\

\midrule

\multirow{5}{*}{\textbf{TPR}}
& \textbf{LMT (ours)}  & $0.94 \pm 0.07$ & $0.89 \pm 0.11$ & $0.65 \pm 0.12$ \\
& CASCADE   & $0.45 \pm 0.13$ & $0.44 \pm 0.02$ & $0.37 \pm 0.15$ \\
& VOC-Emb       & $0.60 \pm 0.21$ & $0.10 \pm 0.02$ & $0.06 \pm 0.04$ \\
& Pure-LLM      & $0.05 \pm 0.06$ & $0.04 \pm 0.06$ & $0.01 \pm 0.02$ \\

\midrule

\multirow{5}{*}{\textbf{FDR}}
& \textbf{LMT (ours)}  & $0.16 \pm 0.08$ & $0.20 \pm 0.13$ & $0.33 \pm 0.19$ \\
& CASCADE   & $0.28 \pm 0.05$ & $0.57 \pm 0.16$ & $0.62 \pm 0.08$ \\
& VOC-Emb        & $0.59 \pm 0.13$ & $0.31 \pm 0.06$ & $0.84 \pm 0.12$ \\
& Pure-LLM      & $0.50\pm 0.07$ & $0.10 \pm 0.22$ & $0.19 \pm 0.32$ \\

\bottomrule
\end{tabular}}
\caption{Comparison of graph recovery metrics (SHD, TPR, and FDR) across different numbers of clusters ($p=5,8,16$) when \textbf{N = 500}. Each value reports the mean performance over 10 runs with different random seeds, together with the standard deviation.}
\label{Tab:event500}
\end{table}

\begin{table}[H]
\centering
\scalebox{0.75}{
\begin{tabular}{c l c c c}
\toprule
\textbf{Metric} & \textbf{Method} & $p=5$ & $p=8$ & $p=16$ \\
\midrule

\multirow{4}{*}{\textbf{SHD}}
& \textbf{LMT (ours)}  & $1.6 \pm 0.9$ & $7.8 \pm 2.2$ & $11.9 \pm 1.1$ \\
& CASCADE   & $7.6 \pm 1.6$ & $12.1 \pm 3.5$ & $21.5 \pm 2.8$ \\
& VOC-Emb   & $9.8 \pm 3.1$ & $13.8 \pm 0.4$ & $23.2 \pm 3.1$ \\
& Pure-LLM      & $7.2 \pm 0.8$ & $12.6 \pm 2.2$ & $21.2\pm 0.4$ \\

\midrule

\multirow{5}{*}{\textbf{TPR}}
& \textbf{LMT (ours)}  & $0.80 \pm 0.07$ & $0.71 \pm 0.11$ & $0.61 \pm 0.04$ \\
& CASCADE   & $0.41 \pm 0.10$ & $0.32 \pm 0.05$ & $0.29 \pm 0.07$ \\
& VOC-Emb       & $0.58 \pm 0.07$ & $0.07 \pm 0.05$ & $0.10 \pm 0.03$ \\
& Pure-LLM     & $0.33 \pm 0.16$ & $0.26 \pm 0.15$ & $0.01 \pm 0.02$ \\

\midrule

\multirow{5}{*}{\textbf{FDR}}
& \textbf{LMT (ours)}  & $0.04 \pm 0.06$ & $0.35 \pm 0.07$ & $0.36 \pm 0.02$ \\
& CASCADE   & $0.37 \pm 0.03$ & $0.55 \pm 0.14$ & $0.60 \pm 0.04$ \\
& VOC-Emb        & $0.56 \pm 0.12$ & $0.30 \pm 0.07$ & $0.35 \pm 0.07$ \\
& Pure-LLM      & $0.27 \pm 0.23$ & $0.32 \pm 0.21$ & $0.13 \pm 0.02$ \\
\bottomrule
\end{tabular}}
\caption{Comparison of graph recovery metrics (SHD, TPR, and FDR) across different numbers of clusters ($p=5,8,16$) when \textbf{N = 1000}. Each value reports the mean performance over 10 runs with different random seeds, together with the standard deviation.}
\label{Tab:event1000}
\end{table}

As discussed above, the simulated dataset is constructed under a chemical engineering system scenario. 
Therefore, the learned causal graph can provide meaningful insights for upstream--downstream process analysis. 
For instance, when $p=5$, the five clusters are designed to correspond to meaningful process units reflected in the textual records: Feed, Reactor, Catalyst, Column, and Recovery/SIS, as summarized in Table~\ref{tab:cluster_stage}. These clusters can be directly grouped according to the keywords in the textual records, since the alarm descriptions explicitly contain process-stage terms. In Figure~\ref{fig:k5_500_event_compare}, each process stage corresponds to one of the five process units, indexed from 1 to 5.


The ground-truth graph represents a multi-branch upstream-to-downstream propagation structure. 
Specifically, Feed-related abnormalities may trigger Reactor, Catalyst, and Column alarms; Reactor abnormalities may further influence Catalyst and Recovery/SIS events; Catalyst-related abnormalities may propagate to Column and Recovery/SIS; and Column abnormalities may finally lead to Recovery/SIS events. Therefore, Recovery/SIS acts as a downstream aggregation node in the simulated process. 

This example illustrates the interpretability of the learned graph at two levels. Each node corresponds to a meaningful process unit, and each directed edge represents a possible temporal propagation relation between process abnormalities. 
For instance, an edge from Feed to Reactor suggests that feed-side alarms tend to precede reactor-side alarms, while an edge from Column to Recovery/SIS suggests that downstream column abnormalities may trigger recovery or safety-interlock events. Such information can support downstream analysis, including alarm prioritization, root-cause tracing, and event propagation analysis.

\begin{table}[H]
\centering

\label{tab:k5_cluster_stage}
\begin{tabular}{c l l}
\toprule
\textbf{Cluster} & \textbf{Process Stage} & \textbf{Interpretation} \\
\midrule
Cluster 1 & Feed & Feed-side abnormalities or feed-related alarms \\
Cluster 2 & Reactor & Reactor operation abnormalities or reactor-related alarms \\
Cluster 3 & Catalyst & Catalyst degradation or catalyst-related alarms \\
Cluster 4 & Column & Distillation-column abnormalities or column-related alarms \\
Cluster 5 & Recovery/SIS & Recovery-unit or safety-interlock-related alarms (SIS) \\
\bottomrule
\end{tabular}
\caption{Process-stage of the five clusters in the $p=5$.}
\label{tab:cluster_stage}
\end{table}

\subsection{Ablation Study}

We conduct an ablation study on the prior to evaluate the contribution of textual information. Specifically, instead of constructing the prior from event descriptions, we replace it with a uniform prior over all possible edges, so that every cluster pair is assigned the same prior edge probability. This setting allows us to examine how much the temporal likelihood alone can recover the causal structure without semantic prior information.

We consider both small-data and large-data settings. For the small-data regime, we test $N=50$ and $N=100$ with $p=5$ and $p=8$. For the larger-scale regime, we test $N=500$ and $N=1000$ with $p=16$. In this way, the ablation covers scenarios ranging from limited numbers of events and clusters to relatively larger event sets and more complex graphs.

\paragraph{Results.}
The ablation results are reported in Tables~\ref{tab:ablation_n50}, \ref{tab:ablation_n100}, and \ref{tab:ablation_p16}. Each value is reported as the mean and standard deviation over 5 repetitions. Overall, LMT consistently outperforms the uniform-prior variant, which confirms the importance of incorporating textual information into the prior.

The advantage of the proposed prior is particularly pronounced in the small-data regime. For example, when $N=50$, LMT reduces SHD from $6.8$ to $2.8$ for $p=5$, and from $14.6$ to $7.3$ for $p=8$. Similar improvements are observed for $N=100$, where SHD decreases from $5.2$ to $2.6$ for $p=5$, and from $14.2$ to $6.0$ for $p=8$. In addition, LMT achieves substantially lower FDR in all small-data settings, indicating that the text-informed prior effectively suppresses spurious edges when temporal evidence is limited.

As the number of events increases, the gap between LMT and the uniform-prior variant becomes smaller, since the temporal likelihood itself becomes more informative with more observed events. Nevertheless, LMT still maintains clear advantages in SHD and TPR for $p=16$ under both $N=500$ and $N=1000$, and also achieves lower FDR for $N=1000$. These results suggest that the text-informed prior is especially beneficial when the available event records are limited, while remaining helpful even in larger-data settings.

Especially when $N\leq 100$, we find that the improvement of LMT over the Uniform-Prior variant is relatively moderate in TPR, but much more pronounced in FDR and SHD. This indicates that the Hawkes likelihood can still identify some true causal edges from temporal information alone, but without an informative prior it is much more prone to retaining spurious edges. Hence, the main benefit of the proposed text-informed prior is to filter out implausible edges early, which is particularly valuable when only limited event records are available.

\textbf{Notably,} evaluating the small-data setting ($N\leq 100$) is particularly important in this work, since it is closer to the real application scenario considered here. In both our simulations and the real engineering alarm-event application discussed later in Section~\ref{sec:RA}, once alarms are triggered and recorded, the manufacturing system is typically stopped and diagnosed after only a short stream of alarm events. In other words, it is often unrealistic for a real engineering system to continue accumulating a large number of alarm-trigger events without intervention. Therefore, unlike more general settings with abundant observations, our goal is to identify causal relations among event clusters from \textbf{limited event records}. This also explains why the gains brought by the text-informed prior in the small-data regime are especially important in practice.

\begin{table}[H]
\centering
\scalebox{0.75}{
\label{tab:ablation_n50}
\begin{tabular}{c l c c}
\toprule
\textbf{Metric} & \textbf{Method} & \(p=5\) & \(p=8\) \\
\midrule

\multirow{2}{*}{\textbf{SHD}}
& \textbf{LMT (ours)}      & $2.8 \pm 1.6$ & $7.3 \pm 1.7$ \\
& Uniform-Prior            & $6.8 \pm 0.8$ & $14.6 \pm 2.4$ \\

\midrule

\multirow{2}{*}{\textbf{TPR}}
& \textbf{LMT (ours)}      & $0.93 \pm 0.07$ & $0.63 \pm 0.09$ \\
& Uniform-Prior            & $0.88 \pm 0.15$ & $0.57 \pm 0.09$ \\

\midrule

\multirow{2}{*}{\textbf{FDR}}
& \textbf{LMT (ours)}      & $0.22 \pm 0.09$ & $0.18 \pm 0.01$ \\
& Uniform-Prior            & $0.45 \pm 0.03$ & $0.52 \pm 0.08$ \\

\bottomrule
\end{tabular}}

\caption{Ablation study comparing LMT with a uniform-prior variant under the small-data setting with $N=50$.}
\label{tab:ablation_n50}
\end{table}

\begin{table}[H]
\centering
\scalebox{0.75}{
\label{tab:ablation_n100}
\begin{tabular}{c l c c}
\toprule
\textbf{Metric} & \textbf{Method} & \(p=5\) & \(p=8\) \\
\midrule

\multirow{2}{*}{\textbf{SHD}}
& \textbf{LMT (ours)}      & $2.6 \pm 0.5$ & $6.0 \pm 2.4$ \\
& Uniform-Prior            & $5.2 \pm 1.8$ & $14.2 \pm 4.2$ \\

\midrule

\multirow{2}{*}{\textbf{TPR}}
& \textbf{LMT (ours)}      & $0.97 \pm 0.06$ & $0.66 \pm 0.16$ \\
& Uniform-Prior            & $0.90 \pm 0.11$ & $0.61 \pm 0.19$ \\

\midrule

\multirow{2}{*}{\textbf{FDR}}
& \textbf{LMT (ours)}      & $0.23 \pm 0.04$ & $0.12 \pm 0.04$ \\
& Uniform-Prior            & $0.37 \pm 0.08$ & $0.50 \pm 0.12$ \\

\bottomrule
\end{tabular}}
\caption{Ablation study comparing LMT with a uniform-prior variant under the small-data setting with $N=100$.}
\label{tab:ablation_n100}
\end{table}

\begin{table}[H]
\centering
\scalebox{0.75}{
\label{tab:ablation_p16}
\begin{tabular}{c l c c}
\toprule
\textbf{Metric} & \textbf{Method} & \(N=500\) & \(N=1000\) \\
\midrule

\multirow{2}{*}{\textbf{SHD}}
& \textbf{LMT (ours)}      & $7.2 \pm 0.45$ & $11.9 \pm 1.09$ \\
& Uniform-Prior            & $10.7 \pm 2.10$ & $16.6 \pm 0.90$ \\

\midrule

\multirow{2}{*}{\textbf{TPR}}
& \textbf{LMT (ours)}      & $0.64 \pm 0.02$ & $0.61 \pm 0.04$ \\
& Uniform-Prior            & $0.58 \pm 0.05$ & $0.52 \pm 0.03$ \\

\midrule

\multirow{2}{*}{\textbf{FDR}}
& \textbf{LMT (ours)}      & $0.22 \pm 0.03$ & $0.36 \pm 0.02$ \\
& Uniform-Prior            & $0.21 \pm 0.04$ & $0.44 \pm 0.01$ \\

\bottomrule
\end{tabular}}
\caption{Ablation study comparing LMT with a uniform-prior variant under the larger-scale setting with \(p=16\).}
\label{tab:ablation_p16}
\end{table}

\section{Real Application}\label{sec:RA}
\subsection{Settings}

\begin{figure}[H]
    \centering
      \scalebox{0.8}{
    \begin{subfigure}[b]{0.5\textwidth}
        \centering
        \includegraphics[width=\textwidth]{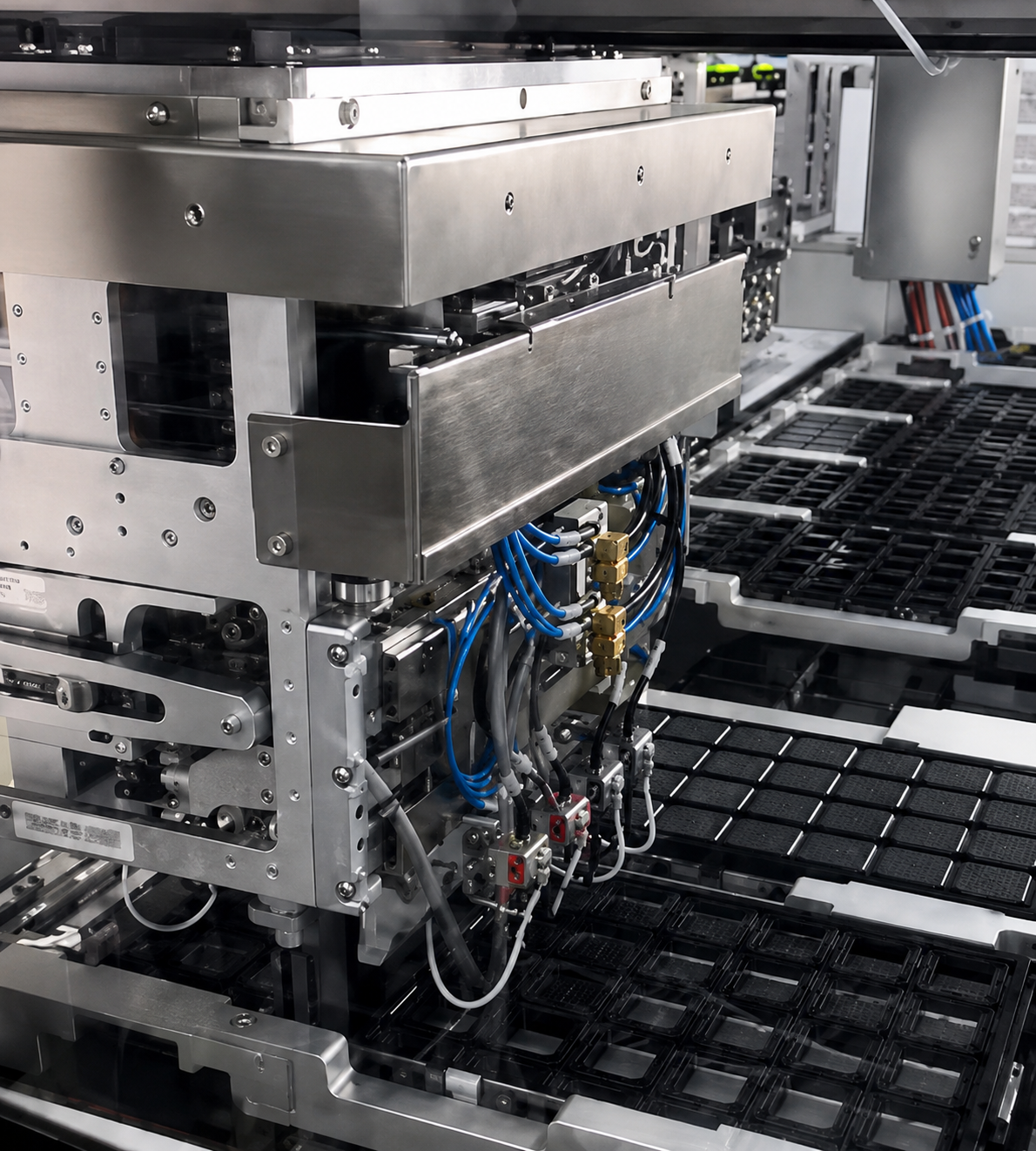}
        \caption{Pick-and-place and tray handling area}
        \label{fig:real_machine_1}
    \end{subfigure}}
    \hfill
      \scalebox{0.8}{\begin{subfigure}[b]{0.5\textwidth}
        \centering
        \includegraphics[width=\textwidth]{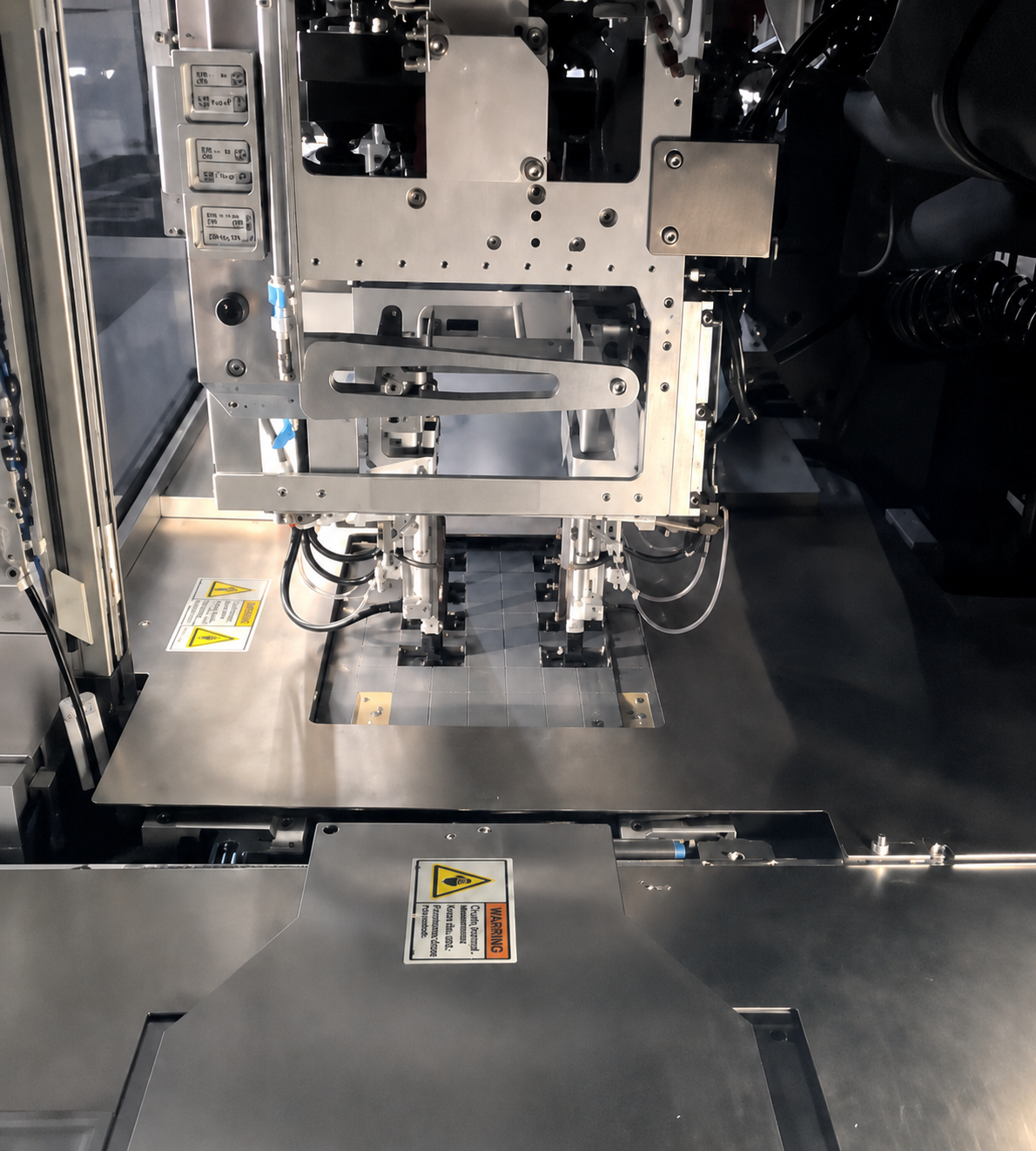}
        \caption{Test-site and positioning module}
        \label{fig:real_machine_2}
    \end{subfigure}}
    \caption{Example images of the  handler. The handler contains multiple mechanical and sensing modules that generate textual alarm records during abnormal operating conditions.}
    \label{fig:real_machine}
\end{figure}

We conduct a case study on alarm-event records collected from a fleet of pick-and-place integrated circuit test handlers deployed at a leading semiconductor backend manufacturing facility. 
These handlers automate the handling of integrated circuit devices during final electrical test: trays of devices are transferred into the handler, picked up and placed into soak chambers for thermal conditioning, transported by shuttles, inserted into test sockets by contactor heads, and finally sorted into output trays based on test results. 
A single handler integrates more than ten functional modules, including the auto tray transfer unit, the input, transfer, and output pick-and-place (PnP) heads, the left and right test-site PnP heads and contactors, the input and output shuttles, the soak chamber, the tray position system, and various pneumatic, vacuum, and electrostatic-discharge subsystems. 
During operation, sensors and controllers across these modules continuously monitor mechanical motion, vacuum levels, pressure, temperature, and timing, and emit textual alarm records whenever an abnormality is detected.
Example alarm records are shown in Table~\ref{tab:alarm-raw}, and example views of the handler are shown in Figure~\ref{fig:real_machine}.
\begin{table}[h]
\centering
\caption{Example records from the production handler alarm dataset.}
\label{tab:alarm-raw}
\small
\begin{tabular}{cll}
\hline
Event id & Alarm Text & Time \\
\hline
1 & Contactor stuck device at right test-site head
  & 2020-05-27 20:24:29 \\
2 & Left input shuttle unable to move off home sensor
  & 2020-05-27 23:14:34 \\
3 & Right test-site PnP unable to pick parts on nests A3, B3
  & 2020-05-29 18:06:21 \\
\hline
\end{tabular}
\end{table}

Our framework is suitable to analyze this dataset.
Specifically, the alarm descriptions are informative: they typically reference a module (``contactor'', ``shuttle'', ``PnP head''), an abnormal behavior (``stuck device'', ``unable to move off home sensor'', ``parts not found on Z-head''), and sometimes a downstream symptom or required operator intervention. 
Such wording carries strong domain-level clues about upstream triggers and downstream consequences that a language model can extract, yet the wording alone is rarely sufficient to settle the direction of influence between two related alarms. 

We select 97 alarm events within a one-week observation window. 
As discussed earlier, the number of alarm events is limited because the handler is often stopped once an alarm occurs, and operators intervene to resolve the problem before the system continues running. 
Therefore, the resulting alarm sequence is short but operationally meaningful. Each alarm event is assigned to one of module-level clusters using the controller field \emph{Source}, which identifies the functional sub-module that raised the alarm (e.g., \texttt{Ts\_Head\_Left}, \texttt{PnP\_Transfer}, or \texttt{Tray\_Xport}).
We apply a deterministic lookup table that maps each observed \emph{Source} tag to a cluster from 1 to 5; several sub-module tags may share the same cluster when they belong to the same process stage (e.g., left and right test-site heads both map to the test cluster). Table~\ref{tab:real_source_cluster_map} gives the complete mapping.

\begin{table}[H]
\centering
  \scalebox{0.75}{
\small
\begin{tabular}{c p{0.42\linewidth} p{0.38\linewidth}}
\toprule
\textbf{Cluster} & \textbf{Observed \emph{Source} tag(s)} & \textbf{Module-level interpretation} \\
\midrule
1 
& \texttt{Ts\_Head\_Left}; \texttt{Ts\_Head\_Right} 
& Test-site PnP heads and contactors \\

2 
& \texttt{PnP\_Transfer} 
& Transfer pick-and-place head \\

3 
& \texttt{PnP\_Input}; \texttt{Shuttle\_Input\_Left} 
& Input-side handling module \\

4 
& \texttt{PnP\_Output} 
& Output pick-and-place and sorting module \\

5 
& \texttt{Tray\_Xport}; \texttt{Tray\_Storage} 
& Tray output, storage, and finally positioning module \\
\bottomrule
\end{tabular}}
\caption{Mapping from controller \emph{Source} tags to five module-level clusters in the handler case study.}
\label{tab:real_source_cluster_map}
\end{table}

\subsection{Results}
\begin{figure}[H]
    \centering
      \scalebox{0.75}{
    \includegraphics[width=0.5\textwidth]{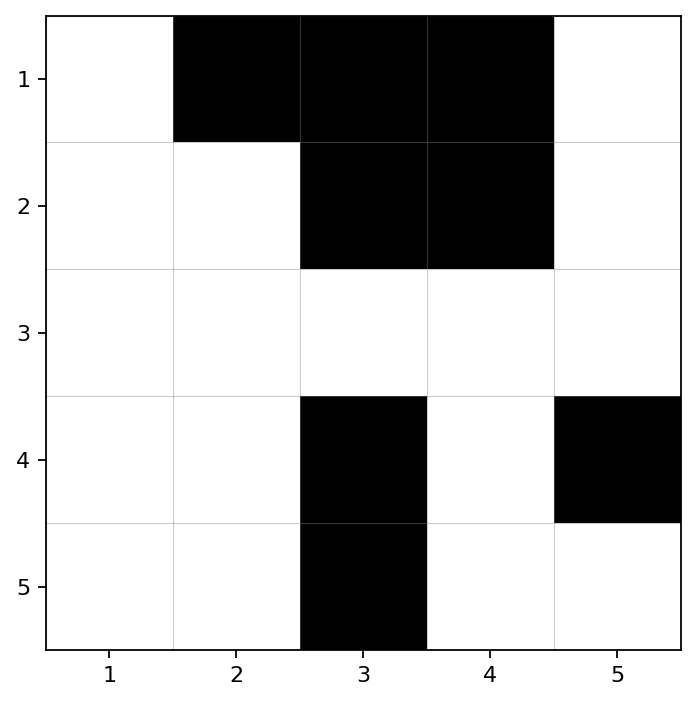}}
    \caption{Learned module-cluster DAG. Each node represents a functional module of the handler, and each directed edge indicates that alarms from the source module tend to precede and potentially trigger alarms in the target module. Black entries denote edges identified in the inferred graph. }
    \label{fig:real_dag}
\end{figure}

The resulting DAG is displayed in Figure~\ref{fig:real_dag}. 
The graph suggests that Test-site/contactor-related alarms act as an upstream alarm source in the observed sequence, with directed edges to the Transfer PnP, Input-side handling, and Output PnP modules. Specifically, test-site failures such as ``stuck device'' or ``lost part'' may cause machine stops, clearing operations, and recovery actions. 
After such interruptions, input-side or transfer-related alarms may appear in the alarm log during the subsequent recovery or restart process. 
Therefore, during real operations, engineers may pay particular attention to the condition of the test-site/contact module, since stabilizing this module may help reduce downstream handler's alarms in later operation cycles.

The graph also indicates that Transfer PnP alarms propagate to both Input-side handling and Output PnP modules, suggesting that pick/place failures can create handling inconsistencies across the system. 
Output PnP alarms further point to Tray transfer/storage alarms, which is consistent with output-side placement or sorting problems leading to tray-level abnormalities. 
In addition, Tray transfer/storage alarms point to Input-side handling alarms, indicating that tray positioning or missing-tray issues may disrupt subsequent input-side feeding operations.

Notably, the Input-side handling module receives incoming edges from multiple other modules in the learned graph.  This pattern suggests that input-side alarms may often appear as recovery-related or restart-related symptoms after abnormalities occur elsewhere in the handler.  For example, alarms from the Test-site/contactor, Transfer PnP, Output PnP, or Tray transfer/storage modules may cause the handler to stop, clear parts, re-home axes, or restart feeding operations, after which input-side alarms can be observed.  Thus, the Input-side handling module should be interpreted as a sensitive module in the alarm-propagation process rather than simply as the first physical stage of material flow.

Overall, the learned graph provides interpretable diagnostic insight by identifying which module-level alarms tend to precede and potentially trigger alarms in other modules.  Such information can support alarm prioritization, root-cause tracing, and preventive-maintenance planning.

Since the true causal graph among alarm clusters is not available in the production setting, the objective of this case study is not quantitative graph recovery against a known ground truth, but rather to demonstrate that our model produces an interpretable cluster-level causal graph whose edges can be inspected and discussed by engineers familiar with handler operation. The learned graph is intended to support root-cause analysis of recurring alarm patterns, alarm prioritization, and preventive-maintenance planning for handlers of this type.

\section{Conclusion}
In this paper, we proposed \textbf{LMT}, a language-model-semantic prior guided Bayesian causal discovery framework with a point-process likelihood derived from event timestamps. The proposed framework is designed to identify causal relationships among clusters of manufacturing process events, particularly for alarm records that contain both textual descriptions and temporal information. Experimental results demonstrate that LMT can effectively obtain causal graph, especially in settings with limited numbers of event records. More broadly, this study illustrates a promising direction for incorporating language-model-derived semantic knowledge into causal analysis of manufacturing event data. Moreover, our work provides new insights into how semantic knowledge extracted from language models can support causal analysis in advanced manufacturing systems. As advanced manufacturing systems become increasingly AI-enabled and highly automated, the growing availability of heterogeneous textual and temporal records creates new opportunities for leveraging language-model-derived semantic knowledge to support causal analysis and decision-making.

\bibliography{mybib}

\begin{thebibliography}{}

\bibitem[\protect\citeauthoryear{Ansari}{Ansari}{2020}]{ansari2020cost}
Ansari, F. (2020).
\newblock Cost-based text understanding to improve maintenance knowledge intelligence in manufacturing enterprises.
\newblock {\em Computers \& Industrial Engineering\/}~{\em 141}, 106319.

\bibitem[\protect\citeauthoryear{Arnold, Liu, and Abe}{Arnold et~al.}{2007}]{arnold2007temporal}
Arnold, A., Y.~Liu, and N.~Abe (2007).
\newblock Temporal causal modeling with graphical granger methods.
\newblock In {\em Proceedings of the 13th ACM SIGKDD international conference on Knowledge discovery and data mining}, pp.\  66--75.

\bibitem[\protect\citeauthoryear{Ban, Chen, Lyu, Wang, Zhu, and Chen}{Ban et~al.}{2025}]{ban2025llm}
Ban, T., L.~Chen, D.~Lyu, X.~Wang, Q.~Zhu, and H.~Chen (2025).
\newblock Llm-driven causal discovery via harmonized prior.
\newblock {\em IEEE Transactions on Knowledge and Data Engineering\/}.

\bibitem[\protect\citeauthoryear{Bhardwaj, Deep, Veeramani, and Zhou}{Bhardwaj et~al.}{2021}]{bhardwaj2021custom}
Bhardwaj, A.~S., A.~Deep, D.~Veeramani, and S.~Zhou (2021).
\newblock A custom word embedding model for clustering of maintenance records.
\newblock {\em IEEE Transactions on Industrial Informatics\/}~{\em 18\/}(2), 816--826.

\bibitem[\protect\citeauthoryear{Bock and Wei{\ss}}{Bock and Wei{\ss}}{2019}]{bock2019proof}
Bock, S. and M.~Wei{\ss} (2019).
\newblock A proof of local convergence for the adam optimizer.
\newblock In {\em 2019 international joint conference on neural networks (IJCNN)}, pp.\  1--8. IEEE.

\bibitem[\protect\citeauthoryear{Cai, Wu, Qiao, Hao, Zhang, and Zhang}{Cai et~al.}{2022}]{cai2022thps}
Cai, R., S.~Wu, J.~Qiao, Z.~Hao, K.~Zhang, and X.~Zhang (2022).
\newblock Thps: Topological hawkes processes for learning causal structure on event sequences.
\newblock {\em IEEE Transactions on Neural Networks and Learning Systems\/}~{\em 35\/}(1), 479--493.

\bibitem[\protect\citeauthoryear{Cajner, Crane, Kurz, Morin, Soto, and Vrankovich}{Cajner et~al.}{2026}]{cajner2026manufacturing}
Cajner, T., L.~D. Crane, C.~Kurz, N.~Morin, P.~E. Soto, and B.~Vrankovich (2026).
\newblock Manufacturing sentiment: forecasting industrial production with text analysis.
\newblock {\em Journal of Applied Econometrics\/}.

\bibitem[\protect\citeauthoryear{Chang, Chen, Cai, and Yu}{Chang et~al.}{2025}]{chang2025anhp}
Chang, Y., D.~Chen, Y.~Cai, and W.~Yu (2025).
\newblock Anhp: Adaptive neural hawkes processes for causal structure learning on event sequences.
\newblock {\em IEEE Internet of Things Journal\/}.

\bibitem[\protect\citeauthoryear{Chen, Carroll, and Morkos}{Chen et~al.}{2023}]{chen2023text}
Chen, C., C.~Carroll, and B.~Morkos (2023).
\newblock From text to images: Linking system requirements to images using joint embedding.
\newblock {\em Proceedings of the Design Society\/}~{\em 3}, 1985--1994.

\bibitem[\protect\citeauthoryear{Chen, Ma, and Yue}{Chen et~al.}{2025}]{chen2025federated}
Chen, J., Y.~Ma, and X.~Yue (2025).
\newblock Federated learning of dynamic bayesian network via continuous optimization from time series data.
\newblock {\em IEEE Transactions on Artificial Intelligence\/}.

\bibitem[\protect\citeauthoryear{Chen, Qin, Jiang, and Choi}{Chen et~al.}{2024}]{chen2024large}
Chen, R., C.~Qin, W.~Jiang, and D.~Choi (2024).
\newblock Is a large language model a good annotator for event extraction?
\newblock In {\em Proceedings of the AAAI conference on artificial intelligence}, Volume~38, pp.\  17772--17780.

\bibitem[\protect\citeauthoryear{Chen, Yan-yi, Tie-zheng, Da-peng, Tao, Zhi, Qing-wen, Hui-han, and Ying-you}{Chen et~al.}{2024}]{chen2024systems}
Chen, W., L.~Yan-yi, G.~Tie-zheng, L.~Da-peng, H.~Tao, L.~Zhi, Y.~Qing-wen, W.~Hui-han, and W.~Ying-you (2024).
\newblock Systems engineering issues for industry applications of large language model.
\newblock {\em Applied Soft Computing\/}~{\em 151}, 111165.

\bibitem[\protect\citeauthoryear{Cheng, Zeng, Hu, Si, and Liu}{Cheng et~al.}{2025}]{cheng2025survey}
Cheng, Q., Z.~Zeng, X.~Hu, Y.~Si, and Z.~Liu (2025).
\newblock A survey of event causality identification: Taxonomy, challenges, assessment, and prospects.
\newblock {\em ACM Computing Surveys\/}~{\em 58\/}(3), 1--37.

\bibitem[\protect\citeauthoryear{Cui, Sun, and Xie}{Cui et~al.}{2023}]{cui2023robust}
Cui, D., Q.~Sun, and M.~Xie (2023).
\newblock Robust statistical modeling of heterogeneity for repairable systems using multivariate gaussian convolution processes.
\newblock {\em IEEE Transactions on Reliability\/}~{\em 72\/}(4), 1493--1506.

\bibitem[\protect\citeauthoryear{C{\"u}ppers, Xu, Musa, and Vreeken}{C{\"u}ppers et~al.}{2024}]{cuppers2024causal}
C{\"u}ppers, J., S.~Xu, A.~Musa, and J.~Vreeken (2024).
\newblock Causal discovery from event sequences by local cause-effect attribution.
\newblock {\em Advances in Neural Information Processing Systems\/}~{\em 37}, 24216--24241.

\bibitem[\protect\citeauthoryear{de~Oliveira, Reis, J{\'u}nior, Cavalcante, de~Lima, Soares, and Marchiori}{de~Oliveira et~al.}{2024}]{de2024impact}
de~Oliveira, A.~H., P.~A. Reis, F.~S. J{\'u}nior, M.~S. Cavalcante, J.~V. de~Lima, L.~F. Soares, and L.~H. Marchiori (2024).
\newblock Impact analysis of using natural language processing and large language model on automated correction of systems engineering requirements.
\newblock In {\em INCOSE International Symposium}, Volume~34, pp.\  992--1007. Wiley Online Library.

\bibitem[\protect\citeauthoryear{Finkbeiner, Frenkel, Metzger, and Siber}{Finkbeiner et~al.}{2024}]{finkbeiner2024synthesis}
Finkbeiner, B., H.~Frenkel, N.~Metzger, and J.~Siber (2024).
\newblock Synthesis of temporal causality.
\newblock In {\em International Conference on Computer Aided Verification}, pp.\  87--111. Springer.

\bibitem[\protect\citeauthoryear{Geigle, Mei, and Zhai}{Geigle et~al.}{2018}]{geigle2018feature}
Geigle, C., Q.~Mei, and C.~Zhai (2018).
\newblock Feature engineering for text data.
\newblock In {\em Feature engineering for machine learning and data analytics}, pp.\  15--54. CRC Press.

\bibitem[\protect\citeauthoryear{Gong, Zhang, Yao, Bi, Li, and Xu}{Gong et~al.}{2024}]{gong2024causal}
Gong, C., C.~Zhang, D.~Yao, J.~Bi, W.~Li, and Y.~Xu (2024).
\newblock Causal discovery from temporal data: An overview and new perspectives.
\newblock {\em ACM Computing Surveys\/}~{\em 57\/}(4), 1--38.

\bibitem[\protect\citeauthoryear{Hurst, Lerer, Goucher, Perelman, Ramesh, Clark, Ostrow, Welihinda, Hayes, Radford, et~al.}{Hurst et~al.}{2024}]{hurst2024gpt}
Hurst, A., A.~Lerer, A.~P. Goucher, A.~Perelman, A.~Ramesh, A.~Clark, A.~Ostrow, A.~Welihinda, A.~Hayes, A.~Radford, et~al. (2024).
\newblock Gpt-4o system card.
\newblock {\em arXiv preprint arXiv:2410.21276\/}.

\bibitem[\protect\citeauthoryear{Jain}{Jain}{2022}]{jain2022hugging}
Jain, S.~M. (2022).
\newblock Hugging face.
\newblock In {\em Introduction to transformers for NLP: With the hugging face library and models to solve problems}, pp.\  51--67. Springer.

\bibitem[\protect\citeauthoryear{Jin, Guo, Chen, Weiskopf, Gotz, and Cao}{Jin et~al.}{2020}]{jin2020visual}
Jin, Z., S.~Guo, N.~Chen, D.~Weiskopf, D.~Gotz, and N.~Cao (2020).
\newblock Visual causality analysis of event sequence data.
\newblock {\em IEEE transactions on visualization and computer graphics\/}~{\em 27\/}(2), 1343--1352.

\bibitem[\protect\citeauthoryear{Kiciman, Ness, Sharma, and Tan}{Kiciman et~al.}{2023}]{kiciman2023causal}
Kiciman, E., R.~Ness, A.~Sharma, and C.~Tan (2023).
\newblock Causal reasoning and large language models: Opening a new frontier for causality.
\newblock {\em Transactions on Machine Learning Research\/}.

\bibitem[\protect\citeauthoryear{Kwon and Lee}{Kwon and Lee}{2025}]{kwon2025predictive}
Kwon, O.-R. and G.~T. Lee (2025).
\newblock A predictive model based on transformer with statistical feature embedding in manufacturing sensor dataset.
\newblock {\em International Journal of Computer Integrated Manufacturing\/}, 1--16.

\bibitem[\protect\citeauthoryear{Li, Han, Liu, Ding, Jing, Zhang, Li, Du, Li, Zhang, et~al.}{Li et~al.}{2025}]{li2025event}
Li, B., X.~Han, J.~Liu, Y.~Ding, L.~Jing, Z.~Zhang, J.~Li, X.~Du, F.~Li, M.~Zhang, et~al. (2025).
\newblock Event extraction in large language model.
\newblock {\em arXiv preprint arXiv:2512.19537\/}.

\bibitem[\protect\citeauthoryear{Li}{Li}{2025}]{li2025poisson}
Li, C.~T. (2025).
\newblock A poisson decomposition for information and the information-event diagram.
\newblock {\em IEEE Transactions on Information Theory\/}.

\bibitem[\protect\citeauthoryear{Li, Shen, Nian, Gao, Wang, Yu, Li, Wang, Hu, and Zhao}{Li et~al.}{2026}]{li2026mitigating}
Li, Y., Y.~Shen, Y.~Nian, J.~Gao, Z.~Wang, C.~Yu, L.~Li, J.~Wang, X.~Hu, and Y.~Zhao (2026).
\newblock Mitigating hallucinations in large language models via causal reasoning.
\newblock In {\em Proceedings of the AAAI Conference on Artificial Intelligence}, Volume~40, pp.\  31852--31860.

\bibitem[\protect\citeauthoryear{Liu, Li, Zhao, Zhang, Meng, Yu, Ji, and Li}{Liu et~al.}{2025}]{liu2025eventgpt}
Liu, S., J.~Li, G.~Zhao, Y.~Zhang, X.~Meng, F.~R. Yu, X.~Ji, and M.~Li (2025).
\newblock Eventgpt: Event stream understanding with multimodal large language models.
\newblock In {\em Proceedings of the Computer Vision and Pattern Recognition Conference}, pp.\  29139--29149.

\bibitem[\protect\citeauthoryear{Liu, Cai, Chen, Qiao, Yan, Li, Zhang, and Hao}{Liu et~al.}{2024}]{liu2024tnpar}
Liu, Y., R.~Cai, W.~Chen, J.~Qiao, Y.~Yan, Z.~Li, K.~Zhang, and Z.~Hao (2024).
\newblock Tnpar: Topological neural poisson auto-regressive model for learning granger causal structure from event sequences.
\newblock In {\em Proceedings of the AAAI Conference on Artificial Intelligence}, Volume~38, pp.\  20491--20499.

\bibitem[\protect\citeauthoryear{Long, Pich{\'e}, Zantedeschi, Schuster, and Drouin}{Long et~al.}{2023}]{long2023causal}
Long, S., A.~Pich{\'e}, V.~Zantedeschi, T.~Schuster, and A.~Drouin (2023).
\newblock Causal discovery with language models as imperfect experts.
\newblock {\em arXiv preprint arXiv:2307.02390\/}.

\bibitem[\protect\citeauthoryear{Lorch, Rothfuss, Sch{\"o}lkopf, and Krause}{Lorch et~al.}{2021}]{lorch2021dibs}
Lorch, L., J.~Rothfuss, B.~Sch{\"o}lkopf, and A.~Krause (2021).
\newblock Dibs: Differentiable bayesian structure learning.
\newblock {\em Advances in Neural Information Processing Systems\/}~{\em 34}, 24111--24123.

\bibitem[\protect\citeauthoryear{Ma, Wang, and Peng}{Ma et~al.}{2023}]{ma2023nonlinear}
Ma, L., M.~Wang, and K.~Peng (2023).
\newblock Nonlinear dynamic granger causality analysis framework for root-cause diagnosis of quality-related faults in manufacturing processes.
\newblock {\em IEEE Transactions on Automation Science and Engineering\/}~{\em 21\/}(3), 3554--3563.

\bibitem[\protect\citeauthoryear{Mbogu and Nicholson}{Mbogu and Nicholson}{2024}]{mbogu2024data}
Mbogu, H.~M. and C.~D. Nicholson (2024).
\newblock Data-driven root cause analysis via causal discovery using time-to-event data.
\newblock {\em Computers \& Industrial Engineering\/}~{\em 190}, 109974.

\bibitem[\protect\citeauthoryear{Mirza}{Mirza}{2021}]{mirza2021event}
Mirza, P. (2021).
\newblock Event causality.
\newblock {\em Computational Analysis of Storylines: Making Sense of Events\/}~{\em 106}, 106--124.

\bibitem[\protect\citeauthoryear{Nadim, Ragab, and Ouali}{Nadim et~al.}{2023}]{nadim2023data}
Nadim, K., A.~Ragab, and M.-S. Ouali (2023).
\newblock Data-driven dynamic causality analysis of industrial systems using interpretable machine learning and process mining.
\newblock {\em Journal of Intelligent Manufacturing\/}~{\em 34\/}(1), 57--83.

\bibitem[\protect\citeauthoryear{Pearl}{Pearl}{2009}]{pearl2009causal}
Pearl, J. (2009).
\newblock Causal inference in statistics: An overview.

\bibitem[\protect\citeauthoryear{Rebboud, Lisena, and Troncy}{Rebboud et~al.}{2022}]{rebboud2022beyond}
Rebboud, Y., P.~Lisena, and R.~Troncy (2022).
\newblock Beyond causality: Representing event relations in knowledge graphs.
\newblock In {\em International conference on knowledge engineering and knowledge management}, pp.\  121--135. Springer.

\bibitem[\protect\citeauthoryear{Salehi, Trouleau, Grossglauser, and Thiran}{Salehi et~al.}{2019}]{salehi2019learning}
Salehi, F., W.~Trouleau, M.~Grossglauser, and P.~Thiran (2019).
\newblock Learning hawkes processes from a handful of events.
\newblock {\em Advances in neural information processing systems\/}~{\em 32}.

\bibitem[\protect\citeauthoryear{Singh, Fry, Perelman, Tart, Ganesh, El-Kishky, McLaughlin, Low, Ostrow, Ananthram, et~al.}{Singh et~al.}{2025}]{singh2025openai}
Singh, A., A.~Fry, A.~Perelman, A.~Tart, A.~Ganesh, A.~El-Kishky, A.~McLaughlin, A.~Low, A.~Ostrow, A.~Ananthram, et~al. (2025).
\newblock Openai gpt-5 system card.
\newblock {\em arXiv preprint arXiv:2601.03267\/}.

\bibitem[\protect\citeauthoryear{Tang, Li, Li, Zhang, Jee, Xiao, Wu, Rhee, Xu, and Li}{Tang et~al.}{2018}]{tang2018nodemerge}
Tang, Y., D.~Li, Z.~Li, M.~Zhang, K.~Jee, X.~Xiao, Z.~Wu, J.~Rhee, F.~Xu, and Q.~Li (2018).
\newblock Nodemerge: Template based efficient data reduction for big-data causality analysis.
\newblock In {\em Proceedings of the 2018 ACM SIGSAC conference on computer and communications security}, pp.\  1324--1337.

\bibitem[\protect\citeauthoryear{Vashishtha, Reddy, Kumar, Bachu, Balasubramanian, and Sharma}{Vashishtha et~al.}{2023}]{vashishtha2023causal}
Vashishtha, A., A.~G. Reddy, A.~Kumar, S.~Bachu, V.~N. Balasubramanian, and A.~Sharma (2023).
\newblock Causal inference using llm-guided discovery.
\newblock {\em arXiv preprint arXiv:2310.15117\/}.

\bibitem[\protect\citeauthoryear{Vukovi{\'c} and Thalmann}{Vukovi{\'c} and Thalmann}{2022}]{vukovic2022causal}
Vukovi{\'c}, M. and S.~Thalmann (2022).
\newblock Causal discovery in manufacturing: A structured literature review.
\newblock {\em Journal of Manufacturing and Materials Processing\/}~{\em 6\/}(1), 10.

\bibitem[\protect\citeauthoryear{Wang, Li, Han, Sarkar, and Zhou}{Wang et~al.}{2017}]{wang2017predictive}
Wang, J., C.~Li, S.~Han, S.~Sarkar, and X.~Zhou (2017).
\newblock Predictive maintenance based on event-log analysis: A case study.
\newblock {\em IBM Journal of Research and Development\/}~{\em 61\/}(1), 11--121.

\bibitem[\protect\citeauthoryear{Wu, Yu, Wu, and Tan}{Wu et~al.}{2025}]{wu2025llm}
Wu, X., K.~Yu, J.~Wu, and K.~C. Tan (2025).
\newblock Llm cannot discover causality, and should be restricted to non-decisional support in causal discovery.
\newblock {\em arXiv preprint arXiv:2506.00844\/}.

\bibitem[\protect\citeauthoryear{Xiao, Alharbi, Zhang, Qin, and Yue}{Xiao et~al.}{2025}]{xiao2025bayesian}
Xiao, X., K.~Alharbi, P.~Zhang, H.~Qin, and X.~Yue (2025).
\newblock Bayesian federated causal inference and its application in manufacturing.
\newblock {\em Journal of Intelligent Manufacturing\/}, 1--25.

\bibitem[\protect\citeauthoryear{Xiao, Shen, and Yue}{Xiao et~al.}{2025}]{xiao2025causality}
Xiao, X., B.~Shen, and X.~Yue (2025).
\newblock Causality-informed anomaly detection in partially observable sensor networks: Moving beyond correlations.
\newblock {\em arXiv preprint arXiv:2507.09742\/}.

\bibitem[\protect\citeauthoryear{Xie and Mu}{Xie and Mu}{2019}]{xie2019distributed}
Xie, Z. and F.~Mu (2019).
\newblock Distributed representation of words in cause and effect spaces.
\newblock In {\em Proceedings of the AAAI Conference on Artificial Intelligence}, Volume~33, pp.\  7330--7337.

\bibitem[\protect\citeauthoryear{Xu, Farajtabar, and Zha}{Xu et~al.}{2016}]{xu2016learning}
Xu, H., M.~Farajtabar, and H.~Zha (2016).
\newblock Learning granger causality for hawkes processes.
\newblock In {\em International conference on machine learning}, pp.\  1717--1726. PMLR.

\bibitem[\protect\citeauthoryear{Zanna and Sano}{Zanna and Sano}{2026}]{zanna2026uncovering}
Zanna, K. and A.~Sano (2026).
\newblock Uncovering bias paths with llm-guided causal discovery: An active learning and dynamic scoring approach.
\newblock In {\em Proceedings of the AAAI Conference on Artificial Intelligence}, Volume~40, pp.\  39567--39575.

\bibitem[\protect\citeauthoryear{Zhang}{Zhang}{2018}]{zhang2018improved}
Zhang, Z. (2018).
\newblock Improved adam optimizer for deep neural networks.
\newblock In {\em 2018 IEEE/ACM 26th international symposium on quality of service (IWQoS)}, pp.\  1--2. Ieee.

\bibitem[\protect\citeauthoryear{Zheng, Aragam, Ravikumar, and Xing}{Zheng et~al.}{2018}]{zheng2018dags}
Zheng, X., B.~Aragam, P.~K. Ravikumar, and E.~P. Xing (2018).
\newblock Dags with no tears: Continuous optimization for structure learning.
\newblock {\em Advances in neural information processing systems\/}~{\em 31}.

\bibitem[\protect\citeauthoryear{Zhou, Hua, Gu, Lu, Peng, Zheng, Shen, and Bao}{Zhou et~al.}{2021}]{zhou2021end}
Zhou, B., B.~Hua, X.~Gu, Y.~Lu, T.~Peng, Y.~Zheng, X.~Shen, and J.~Bao (2021).
\newblock An end-to-end tabular information-oriented causality event evolutionary knowledge graph for manufacturing documents.
\newblock {\em Advanced Engineering Informatics\/}~{\em 50}, 101441.

\bibitem[\protect\citeauthoryear{Zuo, Cao, Chen, Liu, Zhao, Peng, and Chen}{Zuo et~al.}{2021}]{zuo2021learnda}
Zuo, X., P.~Cao, Y.~Chen, K.~Liu, J.~Zhao, W.~Peng, and Y.~Chen (2021).
\newblock Learnda: Learnable knowledge-guided data augmentation for event causality identification.
\newblock In {\em Proceedings of the 59th Annual Meeting of the Association for Computational Linguistics and the 11th International Joint Conference on Natural Language Processing (Volume 1: Long Papers)}, pp.\  3558--3571.

\end{thebibliography}
\end{document}